\documentclass[letterpaper, 10 pt, conference]{ieeeconf} 
\IEEEoverridecommandlockouts
\usepackage{mathrsfs}
\usepackage{color}
\usepackage{amsfonts,amssymb}
\usepackage{amsfonts}
\usepackage{subfigure}
\usepackage{booktabs}

\usepackage{amsmath}
\usepackage{enumerate}
\usepackage{algorithm}
\usepackage{algorithmic}
\usepackage{bm}
\usepackage{makecell}

\usepackage{mathtools}
\usepackage{bbm}
\usepackage{dsfont}
\usepackage{pifont}
\usepackage{amsthm}
\usepackage{cite}

\let\labelindent\relax
\usepackage{enumitem}
\usepackage{balance}
\begin{document}

\title{Moving Target Interception Considering Dynamic Environment}
\author{Chendi Qu$^\dag$, Jianping He$^\dag$, Jialun Li$^\dag$, Chongrong Fang$^\dag$, and Yilin Mo$^\ddag$
	\thanks{
	 $^\dag$: The Dept. of Automation, Shanghai Jiao Tong University, and Key Laboratory of System Control and Information Processing, Ministry of Education of China, Shanghai, China. E-mail address: \{qucd21, jphe, jialunli, crfang\}@sjtu.edu.cn. 
	}
\thanks{
	 $^\ddag$:  Department of Automation and BNRist, Tsinghua University, Beijing, China. E-mail: ylmo@tsinghua.edu.cn.
	}
\thanks{This work was supported by the NSF of China under Grants 61973218 and 62103266.}
}

\maketitle

\begin{abstract}
The interception of moving targets is a widely studied issue. In this paper, we propose an algorithm of intercepting the moving target with a wheeled mobile robot in a dynamic environment. We first predict the future position of the target through polynomial fitting. The algorithm then generates an interception trajectory with path and speed decoupling. We use Hybrid A* search to plan a path and optimize it via gradient decent method. To avoid the dynamic obstacles in the environment, we introduce ST graph for speed planning. The speed curve is represented by piecewise B\'ezier curves for further optimization. Compared with other interception algorithms, we consider a dynamic environment and plan a safety trajectory which satisfies the kinematic characteristics of the wheeled robot while ensuring the accuracy of interception. Simulation illustrates that the algorithm successfully achieves the interception tasks and has high computational efficiency.

\end{abstract}

\vspace{-3pt}
\section{Introduction}
\vspace{-3pt}
Interception is approaching a moving object until the collision happens. As technologies such as perception \cite{shen2013vision}, vision \cite{lee2010geometric}, and planning \cite{mellinger2011minimum} gradually mature, mobile robots have been applied to various fields. %Compared with humans, robots can better adapt to harsh environmental conditions and have better concealment and continuous operation capabilities. 
It has long been a critical topic of robotics to intercept moving targets in a complex environment with mobile robots. The research can be translated not only into military fields such as automatic guidance systems %\cite{pan2010applying} 
and autonomous surveillance, but also into day-to-day application fields such as soccer robot technology %\cite{makarov2019model}
and household life robots.

To intercept a moving target, the first step is to predict the target’s future trajectory accurately. \cite{das2018trajectory} uses Kalman filter to estimate the trajectory of a falling ball in order to catch it with a robot. A long short-term memory (LSTM) neural network is introduced in \cite{altche2017lstm} to solve the problem of accurately predicting vehicles' trajectories on the highway. However Kalman filter prediction requires the system state equation to be known and LSTM requires a large amount of observation as well as the model training in the early stage. 

Once the interception point is calculated, we need to generate a feasible and safety trajectory from current position of the mobile robot to the target point. \cite{ab2020comparative} reviews the commonly used path planning algorithms. Gao et al. \cite{gao2018online} propose a method to generate safe flight corridors by inflating the path and optimize the trajectory with piecewise B\'ezier curves. \cite{zhang2018autonomous} uses Hybrid A* as a warm-starting and solves the autonomous parking path planning as a non-convex optimization problem. %Given the initial and target states of a UAV, \cite{mueller2013computationally} quickly generates the optimal trajectory with a model predictive control like strategy. 
When the environment is dynamic, \cite{cai2020mobile} summarizes the existing robot planning methods. Zhong et al. \cite{zhong2020hybrid} combine A* algorithm with adaptive window approach for global path planning  and real-time tracking in dynamic environments.

Many different approaches have been proposed to deal with the moving targets interception problem depending on different environments and robot kinematic models. \cite{belkhouche2004tracking} focuses on tracking and intercepting a moving object by a wheeled mobile robot. 
\cite{manyam2019optimal} describes an iterative algorithm to find the optimal interception point of a target in a circular motion. A vision-based approach is proposed in \cite{zhang2016vision} which selects the optimal interception path from a set of third order B\'ezier curves. \cite{zhu2013new} presents a novel ant algorithm to quickly plan a path to the interception point considering complex obstacles. 

Nevertheless, to the best of our knowledge, intercepting a moving target in a dynamic environment with a wheeled mobile robot remains an unsolved problem. There are several challenges about this problem. First, the internal control strategy of the moving target is unknown and its real-time position is the only observable information. Second, given the initial and final states, the trajectory planning can be regarded as a constrained optimization problem. Due to the presence of dynamic obstacles in the environment, the problem is non-convex and time-varying, making it difficult to be solved. What's more, the wheeled robot is coupled in the transverse and longitudinal directions, therefore the planned trajectory needs to meet the curvature constraints.

Motivated by above discussions, this paper proposes a novel algorithm for intercepting in dynamic environment via a wheeled robot. We first predict the future trajectory of the moving target based on historical observation data. Then, we decouple trajectory planning into path planning and speed planning to deal with static and dynamic obstacles avoidance, respectively.
The contributions are summarized as follows. 

\iffalse
\begin{figure}[ht]
\setlength{\abovecaptionskip}{0.05cm}
  \centering 
    \includegraphics[width=0.40\textwidth]{pic/flow.png} 
  \caption{Architecture and information flow of the interception process.} 
  \label{flow}
  \vspace{-6pt}
\end{figure}
\fi

\begin{itemize}
\item We propose a novel algorithm of intercepting a moving target with unknown movement motion by a wheeled mobile robot in a dynamic environment. The internal control policy of the target is unknown and the environment exists dynamic obstacles. The algorithm generates a feasible, effective and obstacle-avoidance trajectory for the interception.

\item We use gradient descent method to optimize the interception path for the wheeled mobile robot which is coupled in the transverse and longitudinal directions, adding the curvature constraint to the path.

\item A simulation platform is built and our algorithm is implemented and conducted on a simulation car. Performance study by simulation demonstrates the effectiveness of the proposed method. 
\end{itemize}

The remainder of the paper is organized as follows. 
Section \ref{preliminary} describes the problem of interest. Section \ref{pathplan} shows the algorithms of trajectory prediction and path planning. Section \ref{speedplan} studies how to conduct speed planning for dynamic obstacles avoidance. Simulation results are shown in Section \ref{simulation}, followed by conclusions and future directions in Section \ref{conclusion}.

\vspace{-3pt}
\section{Preliminaries and Problem Formulation}\label{preliminary}
\vspace{-2pt}
\subsection{Kinematic Model of Mobile Robot}
\vspace{-3pt}

Consider a four-wheeled mobile robot $R_M$ with non-holonomic constraint as the interceptor. The kinematic model is given by
\begin{equation}
    \left\{
        \begin{array}{ll}
            \dot{x}=v\cos{\theta} \\
            \dot{y}=v\sin{\theta} \\
            \dot{\theta} = v \frac{\tan{\delta}}{L}  \\
            \dot{\delta}=\omega
        \end{array}
    \right.
    \vspace{-5pt}
\end{equation}
where $x$,\,$y$ show the position coordinates, $\delta$ is the steering angle, $L$ is the distance between the front and rear wheels and $\theta$ is the orientation of the mobile robot. The linear velocity $v$, angular velocity $\omega$ and acceleration $\dot{v}$ have upper and lower bounds respectively. The mobile robot is constrained to only move in the direction of the heading orientation, i.e., we have 
\begin{equation}
\dot{x}\cos{\theta} - \dot{y} \sin{\theta} = 0
\vspace{-5pt}\end{equation}
Let $\mathcal{S}$ be the motion state of the mobile robot, and then the state of the robot at time $t$ is $\mathcal{S}_{t}=[x(t),y(t),\theta(t)] \in \mathcal{S}$.

\vspace{-3pt}
\subsection{Model of Moving Target}
\vspace{-3pt}
Assume that the moving target $R_T$ has smooth motion and bounded changes both in velocity and acceleration. We consider the situation that the target moves under simple linear motions, where the dynamic satisfies 
\begin{equation}
    \textbf{x}_{k+1}=A\textbf{x}_k+B\textbf{u}_k
\vspace{-5pt}\end{equation}
where $\textbf{x}_k = [x_k, y_k, \dot{x}_k, \dot{y}_k]^T$ is the state vector and $\textbf{u}_k = [u_x, u_y, u_{\dot{x}}, u_{\dot{y}}]^T$ is the internal control vector, and
\begin{equation}\nonumber
    A=\begin{pmatrix}
    \Large I_{2 \times 2} & \Delta t \Large I_{2 \times 2}\\
     \text{0}_{2 \times 2} & \Large I_{2 \times 2}
 
    \end{pmatrix},
    B=\begin{pmatrix}
    \text{0}_{2 \times 2} & \text{0}_{2 \times 2} \\
       \text{0}_{2 \times 2} & \Large I_{2 \times 2}
    \end{pmatrix}
\end{equation}
in which $\Delta t$ is the control time interval of the moving target. 

\vspace{-3pt}
\subsection{Problem of Interest}\label{problemof}\vspace{-3pt}
Consider mobile robot $R_M$ intercepting the moving target $R_T$ in a dynamic environment with obstacles. We have following basic assumptions:

\begin{itemize}[leftmargin=*]
\item The initial state $\textbf{x}_0$, control time interval $\Delta t$ and internal control policy $\textbf{u}_k$ of $R_T$ are unknown to $R_M$.

\item The real-time position of $R_T$ can be observed by $R_M$ and recorded as historical data. The observation model at time $t$ is described as:
\begin{equation}
    \textbf{y}_{t}=C\textbf{x}_t+\textbf{e}_t
\vspace{-5pt}\end{equation}
where $\textbf{y}=[\tilde{x}, \tilde{y}]^T$ is the observation of the position, $\textbf{e}=[e_x, e_y]^T$ is the observation error which conforms to normal distribution. We have
\begin{equation}\nonumber
    C=\begin{pmatrix}
    \Large I_{2 \times 2} & \text{0}_{2 \times 2} \\
       \text{0}_{2 \times 2} & \text{0}_{2 \times 2}
    \end{pmatrix},
    \begin{array}{ll}
        e_x \sim \mathcal{N}(0, \sigma_1^2) \\
        e_y \sim \mathcal{N}(0, \sigma_2^2)
    \end{array}
\vspace{-5pt}\end{equation}

\item The global map of the dynamic environment is known by $R_M$. Denote the obstacles in it as $\mathcal{O}(t)$. Each obstacle $\mathcal{O}_i(t) \in \mathcal{O}(t)$ is described as:
\begin{equation}
    \mathcal{O}_i(t) = \left\{z\in \mathbb{R}^2: G_i(t) z\leq b_i(t)\right\}
\vspace{-5pt}\end{equation}
where $G_i(t)$ and $b_i(t)$ are known matrices with respect to time $t$ and $m$ is the total number of the obstacles. 
\end{itemize}

Assume that the mobile robot $R_M$ starts from point $(x_0, y_0)$. Observe the position of the moving target $R_T$ for $L$ times and predict its future position at time $T$, which is also denoted as the interception point $(x_{int}, y_{int})$. Then, we have
\begin{equation}
\mathcal{S}_0=[x_0, y_0, 0],\, \mathcal{S}_T=[x_{int}, y_{int}, \theta_{int}]
\vspace{-5pt}\end{equation}
for $R_M$. We need to generate a feasible and safety trajectory from $\mathcal{S}_0$ to $\mathcal{S}_T$. The trajectory $D_M(t)$ should avoid all the obstacles, which means 
\begin{equation}
    D_M(t) \cap \mathcal{O}(t)=\varnothing, \, \forall t \in [0,T]
\vspace{-5pt}
\end{equation}
Decouple the trajectory planning into path planning and speed planning. The curvature $\kappa$ along the planned path should under the robot's curvature constraint, which is $\kappa \leq \kappa_{max}$.
During the interception process, we are interested in minimizing the energy consumption of $R_M$. Formulate the speed planning along the trajectory into an optimization problem:
\begin{subequations}\vspace{-5pt}
\begin{alignat}{2}
&\min\limits_{v} \quad &&\int_0^T \dot{v}^2 \mathrm{d}t  \\
&s.t. \qquad &&\vert{v}\vert \leq v_{max}, \, \vert{\dot{v}}\vert \leq \dot{v}_{max}  \\
& &&(1), (6), (7) \, \rm{holds} 
\vspace{-5pt}
\end{alignat}
\end{subequations}
where $\dot{v}^2$ is the square of acceleration of $R_M$ and $T$ is the total time cost. Therefore, (8a) represents the energy consumption along the trajectory while (8b) constrains the velocity and acceleration bounds.

\vspace{-5pt}
\section{Trajectory Prediction and Online Path Planning}\label{pathplan}
\vspace{-5pt}
In this section, we first discuss how to predict future trajectory of the moving target. Then, we will propose a online path planning algorithm to avoid the static obstacles.

\subsection{Trajectory Prediction}\vspace{-3pt}
In order to intercept the moving target $R_T$, it is necessary to predict its future position accurately. However, we have no prior knowledge about $R_T$ and its internal control strategy $\textbf{u}_k$ is unknown. In this condition, compared with other motion prediction and trajectory estimation methods, the polynomial fitting is simple, fast and effective.

Polynomial fitting is to find a set of coefficients so that the n-degree polynomial with these coefficients fits the sample points as much as possible. Observe the moving target $L$ times and record its position $p_i \in \mathbb{R}^2, i = 0,1,\ldots, L-1$, and the time $t_i$ of each observation. Denote the real trajectory of the target as $P(t) \subset \mathbb{R}^2$ and it is assumed to be smooth. To fit this trajectory, approximate the $P(t)$ as a n-degree polynomial $\hat{P}(t)$. 
We have 
\begin{equation}
\begin{aligned}
P(t)
&= \begin{pmatrix} P_x(t) \\ P_y(t) \end{pmatrix} 
\approx \begin{pmatrix} a_n t^n + a_{n-1} t^{n-1} + \cdots + a_0 \\
                   b_n t^n + b_{n-1} t^{n-1} + \cdots + b_0 \end{pmatrix}  \\
&= \begin{pmatrix} t^n & t^{n-1} \cdots & 1 \end{pmatrix} 
   \underbrace{\begin{pmatrix} a_n & b_n  \\
                   a_{n-1} & b_{n-1}  \\
                   \vdots & \vdots \\ 
                   a_0 & b_0  \end{pmatrix}}_{\bm{\eta}}
= \hat{P}(t)
\vspace{-5pt}\end{aligned}
\end{equation}
where $\bm{\eta}$ is the parameter to be obtained.
The parameter of $\hat{P}(t)$ is learned by minimizing the sum of squared errors of all the $L$ observations:
\begin{equation}
    \min\limits_{\bm{\eta}} \sum_{i=0}^{L-1} \Vert \hat{P}(t_i)-p_i \Vert^2_2  
\vspace{-5pt}\end{equation}
which is a least squares approximation problem, and it has a unique solution 
\begin{equation}
\bm{\eta} = (H^T H)^{-1} H^T D
\vspace{-5pt}\end{equation}
where $H = \sum_{i=0}^{L-1} T_i^T T_i$ and $D = \sum_{i=0}^{L-1} T_i^T p_i$, in which $T_i = \begin{pmatrix} t_i^n & t_i^{n-1} & \cdots & 1 \end{pmatrix}$. Therefore, after obtaining the polynomial $\hat{P}(t)$, we calculate $\hat{P}(T)$ as the interception point. 

\vspace{-3pt}
\subsection{Online Path Planning with Hybrid A*}\vspace{-3pt}
Once getting the trajectory of the moving target, the interception point $p_{int}$ after time $T$ is predicted. We need to generate a drivable trajectory for the mobile robot $R_M$ from its current position to the target point and meet the specified interception time $T$. In our algorithm, the trajectory planning is divided into path planning and speed planning. 

Use Hybrid A* search to plan a coarse path to the interception point $p_{int}$. Hybrid A* is a graph search based path planning method which can avoid obstacles and satisfy vehicle kinematics \cite{dolgov2008practical}. 
%It inherits the basic idea from A* search such as open and close tables. 
Each node in Hybrid A* is characterized by $(x, y, \theta)$, where $(x, y)$ is the position coordinates of the robot and $\theta$ is the heading angle respectively. The heuristic function in this algorithm is set as the maximum value between the Dubins curve distance and the Euclidean distance to the target point. And in order to speed up the efficiency, the algorithm will directly use the Dubins curve to shot the target point when searching close to the target point. Denote the start point and the target point as $\textbf{s}_0$ and $\textbf{s}_g$, the desired output is a series of waypoints, $\textbf{s}_0, \textbf{s}_1, \cdots, \textbf{s}_n = \textbf{s}_g$, where $\textbf{s}_i = (x_i, y_i, \theta_i)$. 

Notice that Hybrid A* search may not be able to find the optimal path due to the discretization of the search space. The obtained path is relatively rough and needs further smoothing and optimization. However, the search result is basically guaranteed to lie in the neighborhood of the global optimal solution, so the optimal solution can be found through the gradient descent method described in the next subsection.
\vspace{-3pt}
\subsection{Path Smoothing}\vspace{-3pt}
Although the path searched by the Hybrid A* satisfies the kinematics of the robot, it is still relatively rough. Path smoothing can help shorten the length of the path and reduce energy consumption of the robot during the process.
Use gradient descent method to smooth and optimize the interception path. Assume a sequence of waypoints, $\textbf{x}_0, \textbf{x}_1, \cdots, \textbf{x}_n$ are generated by Hybrid A* search. Define $\Delta \textbf{x}_i = \textbf{x}_i - \textbf{x}_{i-1}$, as the displacement vector between two waypoints and
\begin{equation} \nonumber
\begin{small}
\Delta \phi_i = \vert \arctan \frac{\Delta y_{i+1}}{\Delta x_{i+1}} - \arctan \frac{\Delta y_{i}}{\Delta x_{i}} \vert
\end{small}
\vspace{-5pt}\end{equation}
as the steering angle at point $\textbf{x}_i$.
During the interception process, the mobile robot is expected to stay away from obstacles and drive smoothly, while satisfying the curvature limits. Therefore, the objective function includes collision-avoid term, curvature term and smoothness term, which is formulated as
\begin{equation}
J_G = \omega_o J_{obs} + \omega_{\kappa} J_{cur} + \omega_s J_{smo}
\vspace{-5pt}\end{equation}
where  $\omega_o, \omega_{\kappa}, \omega_s$ are weights using to control the impact of each term on the path respectively, and $\textbf{x}_i (i = 0,\ldots,n)$ are optimization variables.

The first term $J_{obs}$ is to penalize the collision with obstacles. Denote $\textbf{o}_i$ as the nearest obstacle of the current point $\textbf{x}_i$. If the distance between $\textbf{x}_i$ and $\textbf{o}_i$ is less than $d_{max}$, $J_{obs}$ will work as a penalty term.
\vspace{-5pt}\begin{equation}
J_{obs} = \sum_{i=1}^N \sigma_o(\vert \textbf{x}_i - \textbf{o}_i \vert - d_{max})
\vspace{-5pt}
\end{equation}
where $\sigma_o$ is a quadratic penalty function. %The gradient with respect to $\textbf{x}_i$ is calculated as
\iffalse
\begin{equation} \nonumber
\begin{small}
\frac{\partial \sigma_o}{\partial \textbf{x}_i} = 2(\vert \textbf{x}_i -\textbf{o}_i \vert - d_{max} ) \frac{\textbf{x}_i - \textbf{o}_i}{\vert \textbf{x}_i - \textbf{o}_i \vert}
\end{small}
\end{equation}
\fi

The second term $J_{cur}$ is used to limit the curvature of the path below the maximum curvature $\kappa_{max}$ and ensure the drivability of the path. The curvature at point $\textbf{x}_i$ can be approximately calculate as $\kappa_i = \frac{\Delta \phi_i}{\vert \Delta \textbf{x}_i \vert}$. If the curvature exceeds $\kappa_{max}$, the path needs to be adjusted.
\begin{equation}
J_{cur} = \sum_{i=1}^{N-1}  \left( \frac{\Delta \phi_i}{\vert \Delta \textbf{x}_i \vert} - \kappa_{max} \right)
\vspace{-5pt}\end{equation}

The last term $J_{smo}$ evaluates the changes of the displacement vectors in order to smooth the path. $J_{smo}$ and its derivative with respect to $\textbf{x}_i$ are expressed as:
\begin{equation}
J_{smo} = \sum_{i=1}^{N-1} (\Delta \textbf{x}_{i+1} - \Delta \textbf{x}_i )^2
\vspace{-5pt}
\end{equation}
%and
\iffalse
\begin{equation}\nonumber
\begin{small}
\frac{\partial J_{smo}}{\partial \textbf{x}_i} = 2(\textbf{x}_{i+2} - 4\textbf{x}_{i+1} + 6 \textbf{x}_{i} -4 \textbf{x}_{i-1} + \textbf{x}_{i-2})
\end{small}
\end{equation}
\fi

All of the above gradients can be calculated directly and efficiently. Let $\bm{\alpha}$ be the step size of the gradient descent. The iteration formula of $\textbf{x}_k$ is given by:
\begin{equation}
\textbf{x}_{k+1} = \textbf{x}_{k} -( \bm{\alpha}_1 \frac{\partial J_{obs}}{\partial \textbf{x}_k} + \bm{\alpha}_2 \frac{\partial J_{cur}}{\partial \textbf{x}_k} + \bm{\alpha}_3 \frac{\partial J_{smo}}{\partial \textbf{x}_k})
\vspace{-5pt}
\end{equation}
and
\begin{equation}\nonumber
\lim_{k \to \infty} \textbf{x}_k = \textbf{x}^* = \arg\min J_{G}(\textbf{x})
\vspace{-5pt}\end{equation}
As the number of iterations increases, the path gradually converges to the optimal solution $\textbf{x}^*$. The optimized path has shorter interception distance and meets the curvature limit of the mobile robot.

\section{\vspace{-4pt} Online Speed Planning for Dynamic Obstacles Avoidance}\label{speedplan}
After the path planning, the mobile robot is able to avoid static obstacles. In this section we describe the speed planning method, which further reduces the energy consumption and helps with dynamic obstacle avoidance during the interception. 

\subsection{Dynamic Programming with ST Graph}
The speed planning is to append a speed component to the path. We introduce ST graph to analyze this issue \cite{fan2018baidu}. With the help of ST graph, we can plan the state of robot $R_M$ during the interception process to avoid dynamic obstacles and use piecewise B\'ezier curves to optimize the speed. 

In ST graph, the abscissa `t' represents time and the ordinate `s' is the longitudinal distance traveled by the robot. To simplify the problem, consider that the dynamic obstacles are all moving at constant speeds. We project the dynamic obstacles as parallelograms on the ST graph. %The side length of the parallelogram along `s' axis is set as the length of the obstacle plus half the length of the robot. 
Then we need to find a station-time curve for $R_M$ from $(0,0)$ to $(T,s_m)$, where $s_m$ is the total interception distance. For the safety during driving, the curve should not overlap with obstacle areas.

Discretize the ST graph and denote each node as $(t_i, s_i)$. ${\rm cost}(t_i, s_i)$ represents the total cost from $(0, 0)$ to $(t_i, s_i)$, which can be calculated as:
\begin{equation}
\begin{aligned}
& {\rm cost}(t_i, s_j) = \min \limits_{t_i, s_j} 
(
{\rm cost} (t_i, s_j), {\rm cost}(t_{i-1}, s_k) + \\ 
&  {\rm cost}_{node}(t_i, s_j) + {\rm cost}_{edge}((t_{i-1}, s_k),(t_i,s_j))
)
\vspace{-5pt}\end{aligned}
\end{equation}
where ${\rm cost}_{node}(t_i, s_j)$ is the cost of one single node $(t_i, s_i)$ and ${\rm cost}_{edge}((t_{i-1}, s_k),(t_i,s_j))$ is the transition cost between nodes $(t_{i-1}, s_k)$ and $(t_i,s_j)$. After calculating the cost of each node, we use dynamic programming (DP) to find a solution with the least cost as the reference line.

The station-time curve generated by the DP is not smooth and needs to be further optimized. We formulate the curve into piecewise B\'ezier curves with Bernstein polynomial basis as it changes the optimization problem into a quadratic programming problem while ensuring the obstacle-avoidance.

\subsection{Speed Curve Formulation by Piecewise B\'ezier Curves}
The $j$-th piece $n$-th ordered B\'ezier curve is written as:
\begin{equation}\nonumber
B_j(t) = c_j^0 b_n^0(t) + c_j^1 b_n^1(t) + \cdots + c_j^n b_n^n(t) = \sum_{i=0}^n c_j^i b_n^i(t)
\vspace{-5pt}\end{equation}
where Bernstein basis is defined as $b_n^i(t)  = C_n^i \cdot t^i \cdot (1-t)^{n-i} , t\in[0,1]$. The coefficients $c_j = (c_j^0, c_j^1, \cdots, c_j^n)$ are called control points in $j$-th piece. Compared to monomial basis polynomial, B\'ezier curve has following properties:
\begin{itemize}
\item The B\'ezier curve always starts at the first control point $B_j(0) = c_j^0$ and ends at the last point $B_j(1) = c_j^n$.
%\item The variable t is defined on $t\in[0,1]$.
\item The B\'ezier curve is confined within the convex hull of control points.
\item The derivative curve $B^{(1)}(t)$ of the B\'ezier curve is also a B\'ezier polynomial with the control points $c_i^{(1)} = n \cdot (c_{i+1} - c_i)$. In this way, we can calculate arbitrary derivatives of $B(t)$.
\end{itemize}

Thanks to the second property, we can add constraints to the control points and limit the the B\'ezier curve to a safe area without obstacles. We use piecewise polynomial instead of higher order polynomial to guarantee fitting performance while avoiding numerical instability. The station with time can be formulated as
\begin{equation}
s(t) = 
\left\{
\begin{array}{cc}
h_0 B_0(\frac{t-T_0}{h_0}), &t\in[T_0,T_1] \\
h_1 B_1(\frac{t-T_1}{h_1}), &t\in[T_1,T_2] \\
     \cdots \\
h_m B_m(\frac{t-T_m}{h_m}), &t\in[T_m,T_{m+1}]
\end{array}
\right.
\vspace{-5pt}\end{equation}
where $h_j, j = 0,1,\cdots,m$, is used to scale the parameter t to the fixed interval $[0,1]$.

\subsection{Speed Curve Optimization Formulation}
\begin{figure}[htbp]
\setlength{\abovecaptionskip}{0.01cm}
  \centering 
    \includegraphics[width=0.2\textwidth]{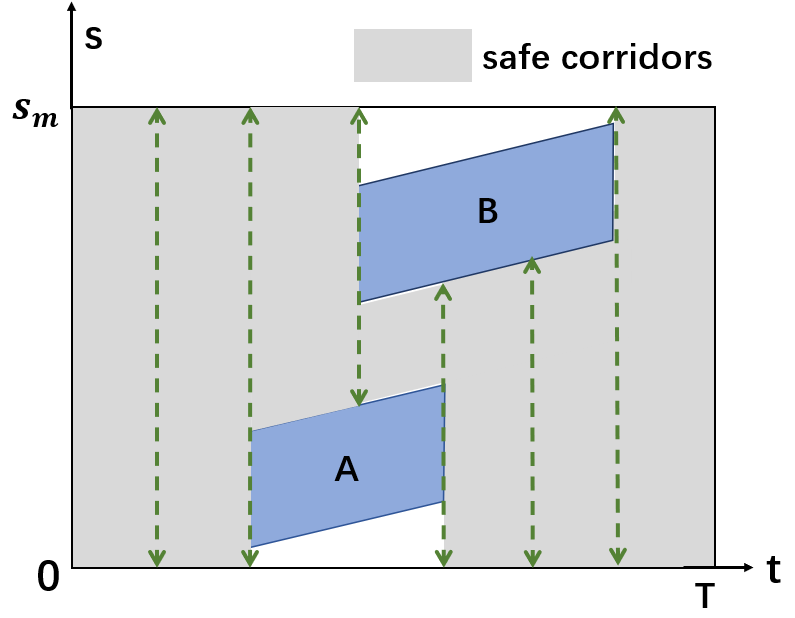} 
  \caption{Safety constraints on ST graph. A and B are projections of dynamic obstacles. Grey trapezoidal areas are generated safe corridors.} 
  \label{stgraph}
  \vspace{-10pt}
\end{figure}

There are two main costs need to be minimized. The optimization object is designed as.
\begin{equation}
J_s = \omega_1 \int_0^T \Ddot{s}(t)^2 \mathrm{d} t + \omega_2 (s(T) - s^\tau(T))^2
\vspace{-5pt}
\end{equation}
where $s^\tau(t)$ is the reference curve obtained by DP in the previous step. The first term is the square of the acceleration, which means to minimize the energy consumption of the mobile robot during the interception process. The second term is to reduce the distance between the last position of the curve and the expected end state $s^\tau(T)$.
The followings are constraints.

 i) Waypoints constraints: Each piecewise curve starts from the certain position, speed and acceleration.
\begin{equation}
h_j^{(1-l)} \cdot c_j^{l,0} = \frac{\mathrm{d}^l s}{\mathrm{d}t^l} \bigg|_{t=0}, l = 0,1,2
\vspace{-5pt}
\end{equation}
where $c_j^{l,i-1}$ is the $i$-th control points for the $l$-th derivative of $j$-th B\'ezier curve.

 ii) Continuity constraints: The position, speed and acceleration are continuous at the connection of each segment curve.
\begin{equation}
h_j^{(1-l)} \cdot c_j^{l,n} = h_{j+1}^{(1-l)} \cdot c_{j+1}^{l,0} , l = 0,1,2, j = 0, \cdots, m-1
\vspace{-5pt}\end{equation}

 iii) Safety Constraints: Using the convex hull property of B\'ezier curve, the piecewise curve is ensured to be collision-free through limiting the control points in a safe area. As the projections of dynamic obstacles on the ST graph are parallelograms, the feasible area is divided into multiple trapezoidal corridors illustrated in Fig.~\ref{stgraph}. The algorithm of generating trapezoidal corridors in ST graph is given in \cite{li2021speed}. 
 Constrain the control points in the corresponding trapezoidal corridor:
\begin{equation}
\underline{q_j^0} + h_j \underline{q_j^1} M_{i,1} \leq c_j^{0,i} \leq \overline{p_j^0} + h_j \overline{p_j^1} M_{i,1}, j = 0,1,\ldots,m
\vspace{-5pt}\end{equation}

 iv) Dynamical Feasibility Constraints: Use the fourth property of B\'ezier curve to limit the upper and lower bounds of the mobile robot. 
 \vspace{-5pt}
\begin{equation}
\begin{aligned}
&\underline{v_j} \leq c_j^{1,i} \leq \overline{v_j} , \,
\overline{v_j} = \sqrt{\frac{a_{cm}}{\kappa_j}}  \\
&\underline{a_j} \leq c_j^{2,i} \leq \overline{a_j}
\vspace{-5pt}
\end{aligned}
\end{equation}
where $\underline{v_j}, \overline{v_j}$ and $\underline{a_j}, \overline{a_j}$ are lower and upper bounds of the velocity and acceleration of $j-$th curve.

Therefore, the optimization problem (19)-(23) can be formulated as a quadratic programming problem:
\begin{equation}
\begin{split}
& \min \limits_{\boldsymbol{c}} \quad \boldsymbol{c}^T \boldsymbol{Q} \boldsymbol{c} + \boldsymbol{q_{\boldsymbol{c}}}^T \boldsymbol{c} + \rm{const} \\
& \begin{aligned}
s.t.  \qquad
& \boldsymbol{A}_{eq} \boldsymbol{c} = \boldsymbol{b}_{eq}, \,
 \boldsymbol{A}_{iq} \boldsymbol{c} \leq \boldsymbol{b}_{iq}
\vspace{-5pt}\end{aligned}
\end{split}
\end{equation}
%We refer readers to the appendix for the detailed formulation process. 
The detailed formulation process is in \cite{li2021speed}.
This problem can be solved by solvers like OSQP. Notice that (19) is equivalent to (8a), so that the optimization problem (8) is solved.

% \begin{algorithm}[htb]
%  \caption{The Moving Target Interception Algorithm}
%  \begin{algorithmic}
%     \REQUIRE  
%       The static and dynamic obstacles, $\mathcal{O}(t)$; 
%       The initial positions of the mobile robot and the moving target, $[p_0^m, p_0^t]$;
%       The longest distance to determine a successful interception, $\epsilon$;
%     \ENSURE
%       The optimal control strategy of the mobile robot, $\textbf{u}$;  
%     \STATE $dis = \Vert {p_0^m-p_0^t}\Vert_2$ 
%     \WHILE{ $dis > \epsilon$ } 
%     \STATE Observe and record the position of the target for L times, $p_i,i = 0,1,\cdots, L-1$;
%     \STATE Predict and calculate the interception point $p_{int}$ by polynomial fitting (10)-(11);
%     \STATE Update the position of the mobile robot $p_0^m$;
%     \STATE Search a coarse path $\textbf{s}$ from $p_0^m$ to $p_{int}$ with Hybrid A* algorithm;
%     \STATE Solve $\textbf{x}^* = \arg\min J_G$ by gradient decent method using (13)-(16);
%     \STATE Use ST graph to plan a coarse speed curve $s(t)$;
%     \STATE Optimize the speed curve $s(t)$ with piecewise B\'ezier curves (24).
%     \ENDWHILE
%  \RETURN $\textbf{u} = (\textbf{x}, \textbf{v})$.
% \end{algorithmic} 
% \end{algorithm} 

\vspace{-5pt}
\section{ Simulation}\label{simulation}\vspace{-3pt}
In this section, we conduct extensive simulations on each part of our algorithm to demonstrate the feasibility and effectiveness of the interception process. We also build a simulation platform with ROS and Gazebo and implement our system on it in C++11.
\vspace{-3pt}
\subsection{Trajectory Prediction}\vspace{-3pt}
The simulation system omits the perception module in the real environment and directly obtains the real-time position of the mobile robot and the moving target through the simulator. Set the number of observation as $L=15$ and the time interval is $\Delta t = 1s$. Test the trajectory prediction effect of the polynomial fitting when the target moves in two motion modes, which is uniform linear motion with $\dot{x} = 0, \, \dot{y} = 0$ and curve motion with $\dot{x} = 0, \dot{y} = 0.4$ separately. In order to simulate the observation error in the real environment, add Gaussian errors $e_x, e_y \sim \mathcal{N}(0, 0.01)$ to each point. 
\vspace{-5pt}
\begin{figure}[ht]
\setlength{\abovecaptionskip}{0.1cm}
  \centering 
    \subfigure{ 
    \label{pre0} 
    \includegraphics[width=0.22\textwidth]{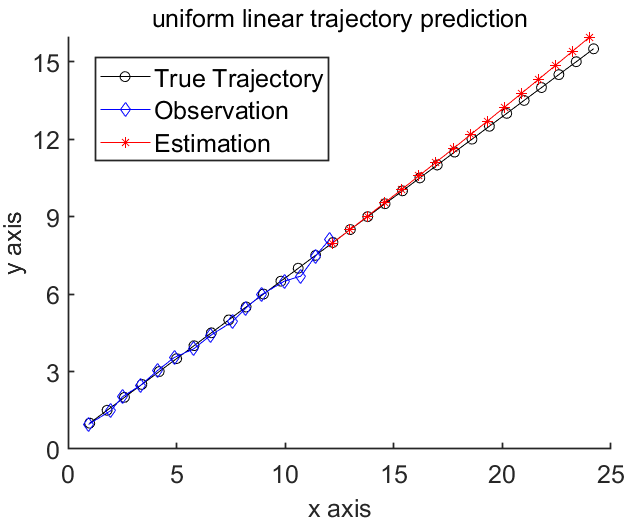} 
    }
    \subfigure{ 
    \label{pre1} 
    \includegraphics[width=0.22\textwidth]{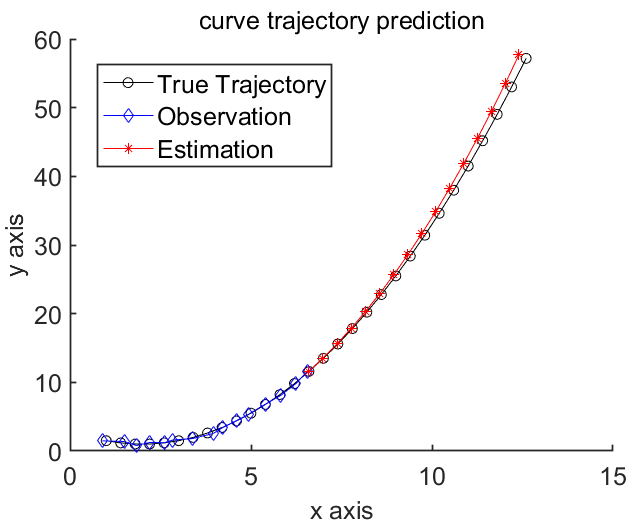} 
    }
  \caption{Trajectory prediction results.} 
  \label{tra_pre}
  \vspace{-10pt}
\end{figure}

The simulation of trajectory prediction results are shown in Fig.~\ref{tra_pre}. The predicted time length is 15s. The black curve is the true trajectory of the target from 0s to 15s, while the blue points are observed positions $p_i$ with observation errors and the red curve is the predicted trajectory. The prediction errors at different time are shown in Tab.~\ref{tab1}, which are relative errors obtained by averaging multiple experiments. We set the time length of a single prediction as 10s, which means $T = 10s$.

\begin{table}
    \centering
    \caption{Average error of trajectory prediction}
    \vspace{-5pt}
	\label{tab1}
    \begin{tabular}{ccccc}
    \toprule    
    Predict Time & 0s & 5s & 10s & 15s  \\    
    \midrule   
    Uniform Linear & 0.23\% & 0.56\% & 1.28\% & 2.04\% \\
    Curve  & 0.58\% & 2.13\% & 3.94\% & 5.88\%  \\
    \bottomrule   
    \end{tabular}
    \vspace{-8pt}
\end{table}
\vspace{-3pt}
\subsection{Path Planning and Optimization}
\vspace{-3pt}
\begin{figure*}[ht]
\setlength{\abovecaptionskip}{0.1cm}
  \centering 
    \subfigure[trajectory prediction t = 0]{ 
    \label{se0}
    \includegraphics[width=0.2\textwidth]{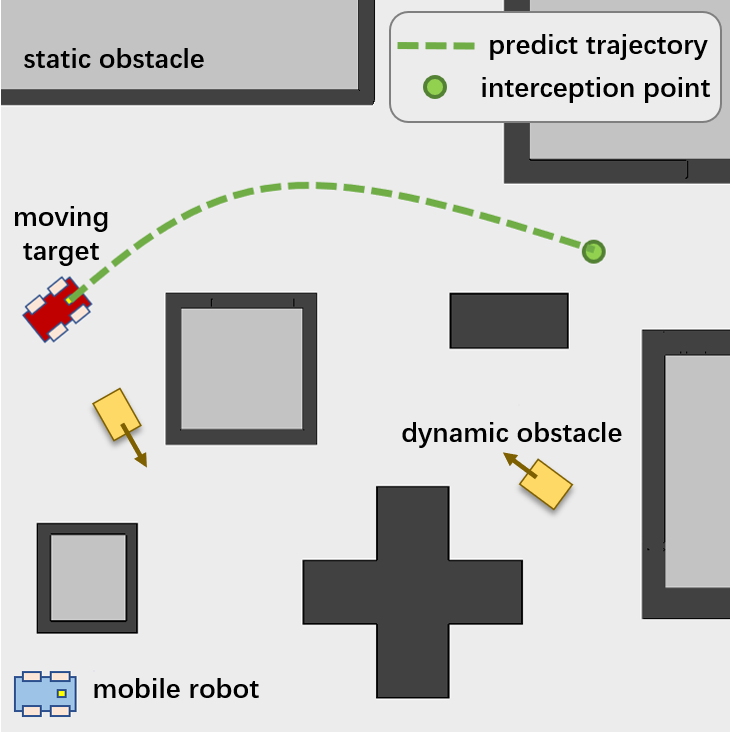} 
    }\hspace{12pt}
    \subfigure[path planning t = 0]{
    \label{se1} 
    \includegraphics[width=0.2\textwidth]{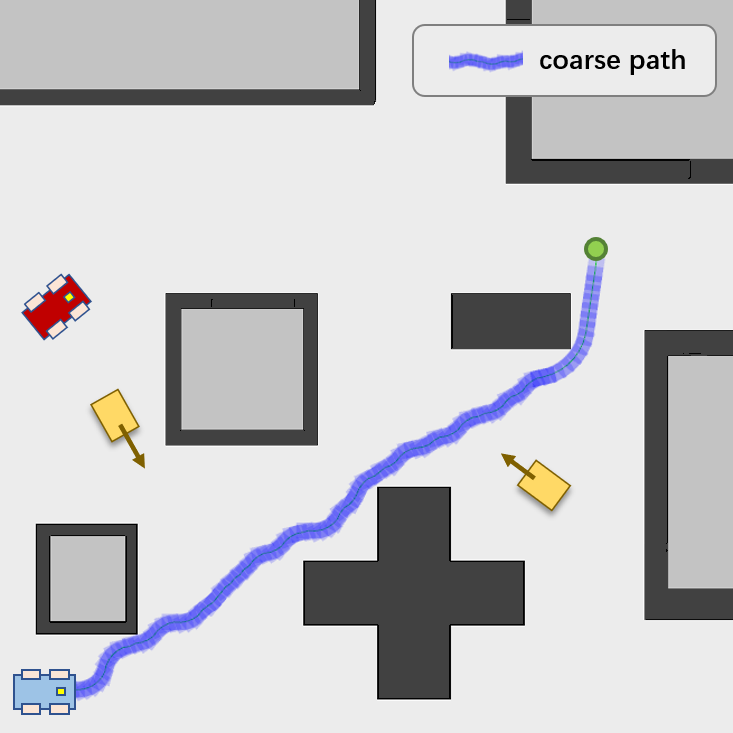} 
    }\hspace{12pt}
    \subfigure[moving 0 $<$ t $<$ T]{ 
    \label{se2} 
    \includegraphics[width=0.2\textwidth]{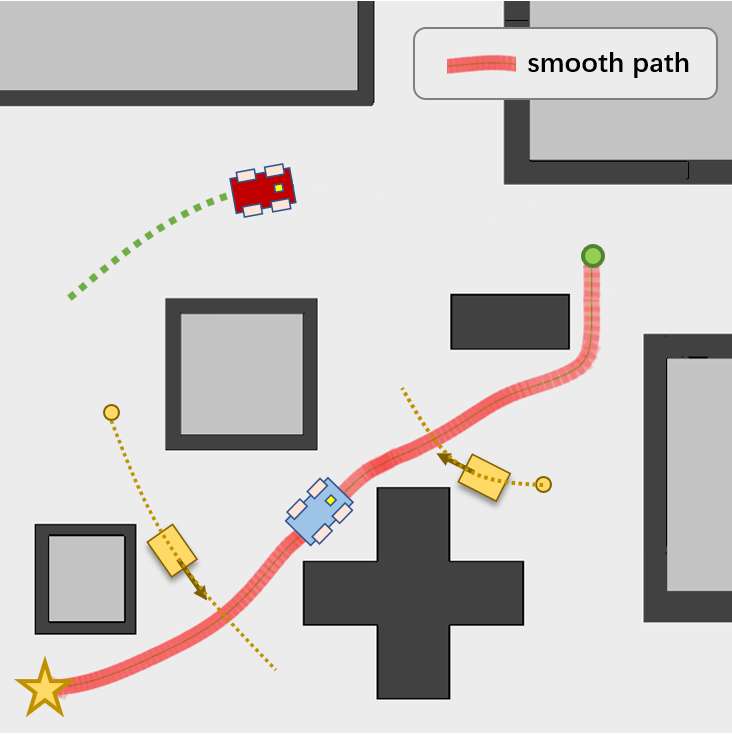}
    }\hspace{12pt}
    \subfigure[successfully intercept t = T]{ 
    \label{se3} 
    \includegraphics[width=0.2\textwidth]{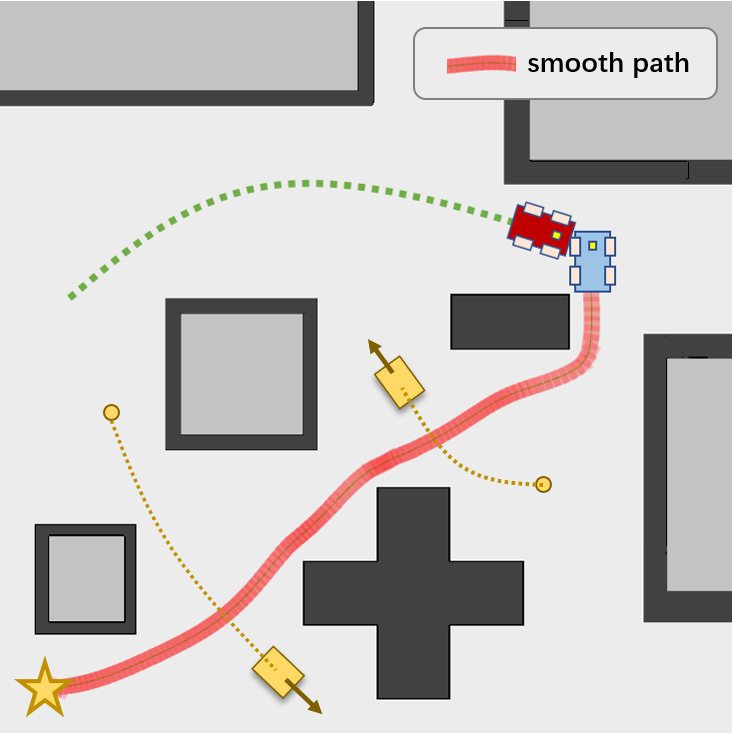} 
    }
  \caption{Path planning results and the illustration of interception process. (a). The red car is the moving target $R_T$ and blue car is the controlled mobile robot $R_M$. The yellow and gray areas are dynamic and static obstacles respectively. The green curve is the predicted trajectory of $R_T$ while the green point is the interception point. (b). The blue curve is the coarse path searched by Hybrid A*. (c). The red curve is the smoothed and optimized path. During the interception, $R_M$ accelerates past the first dynamic obstacle and slows down to avoid the second obstacle. (d). Interception successfully achieved.}
  \label{path_plan}
  \vspace{-10pt}
\end{figure*}

The global environment and obstacles are already known as the input. The size of the map is $80 \times 80$ px. Given the initial state $\mathcal{S}_0=[x_0, y_0, 0]$ and final state $\mathcal{S}_T=[x_{int}, y_{int}, \theta_{int}]$, the algorithm generates an optimized path for the mobile robot $R_M$. Set $\omega_{\sigma} = 0.1, \omega_{\kappa} = 0.1, \omega_s = 0.2$ and the step size $\alpha = 0.25$. Use ROS RViz to visualize the planning results. Simulation results are shown in Fig.~\ref{path_plan}. The blue curve in Fig.~\ref{se1} is the coarse path planned by Hybrid A* search while the red curve in Fig.~\ref{se2} and \ref{se3} is the optimized path obtained by gradient descent method. Compared with the result obtained by using Hybrid A* alone, the optimized path is much smoother and when the searched path is too close to the obstacle, the optimization algorithm adjusts the path to stay away from the obstacle properly.
\vspace{-3pt}
\subsection{Speed Planning}
\vspace{-3pt}
Conduct the dynamic programming algorithm on the ST graph. Assume the interception time is 10s and the total distance of planned path is 6m. Then the initial state of mobile robot $R_M$ is $(t = 0,s = 0)$ and the target state is $(t= 10, s = 6)$. We project the dynamic obstacles which affect $R_M$ during $t = 0 \sim 10s$ on the ST graph. Suppose the initial speed of $R_M$ is $v(0)= 0.5 m/s$ and acceleration is $a(0) = 0 m/s^2$. Set the parameter of the optimization object as $\omega_1 = 10.0,\,\omega_2 = 3.0$. 

\begin{figure*}[t]
  \centering 
  \setlength{\abovecaptionskip}{0.1cm}
  \subfigure[speed planning]
    {
    \includegraphics[width=0.26 \textwidth]{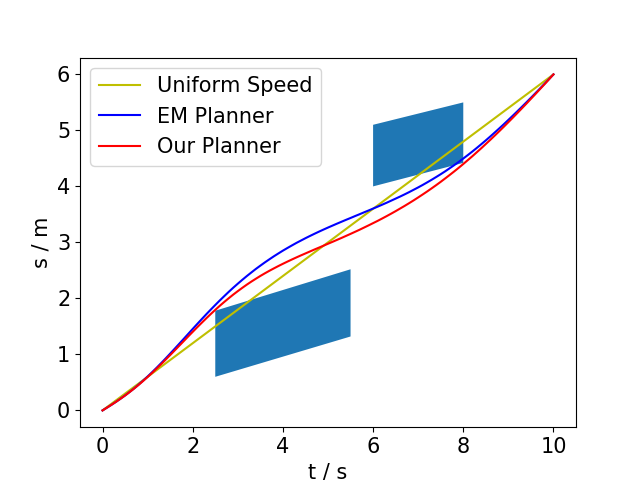} 
    \label{bezier_curve}
    }
    \subfigure[speed curve]
    { 
    \label{speed} %% label for first subfigure 
    \includegraphics[width=0.26\textwidth]{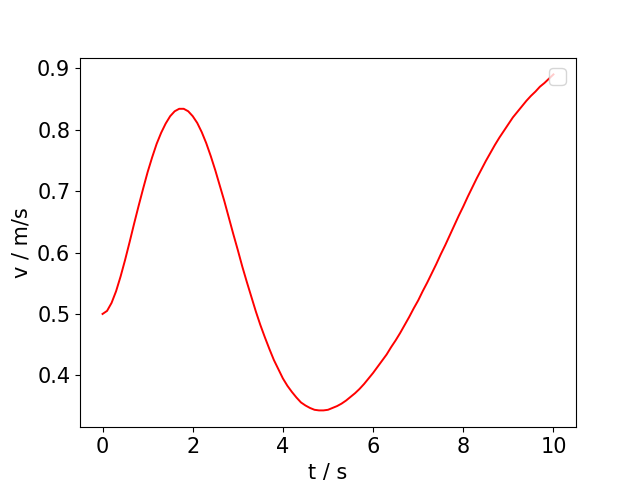} 
    }
    \subfigure[acceleration curve]{ 
    \label{acc} %% label for first subfigure 
    \includegraphics[width=0.26\textwidth]{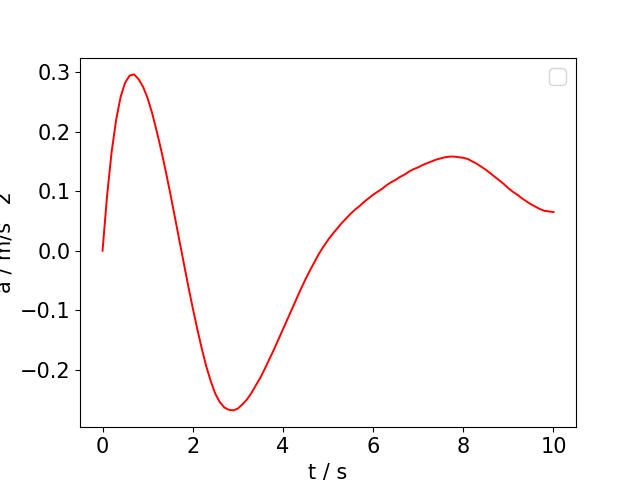} 
    }
  \caption{Speed planning by piecewise B\'ezier curves results. (a). The uniform speed (yellow line) and speed planned by EM planner (blue line) both hit dynamic obstacles, while the speed curve planned by our planner (red line) avoids dynamic obstacles and ensures the minimum energy consumption. (b)(c). The speed and acceleration curves are smooth and easy to follow.}
  \label{s_acc}
  \vspace{-10pt}
\end{figure*}

As shown in Fig.~\ref{bezier_curve}, piecewise B\'ezier curves are calculated and optimized . Fig.~\ref{speed} illustrates corresponding speed profiles and Fig.~\ref{acc} shows acceleration profiles. We use average acceleration $\sqrt{\frac{1}{T} \int_0^T (\Ddot{s}(t))^2 {\rm d} t}$ to quantify the energy consumption during the interception. The comparison between our algorithm and the EM 
Motion planner \cite{fan2018baidu}, which is one of the start-of-art methods and has been used in the autonomous driving projects of Baidu Inc., in the acceleration and computing time is shown in Tab.~\ref{tab2}.

\begin{table}
    \centering
    \caption{speed planner comparison}
    \vspace{-5pt}
	\label{tab2}
    \begin{tabular}{ccccc}
    \toprule    
    Speed Planner & Max Acc. & Ave. Acc. & Ave. Time Cost\\ 
    \midrule   
    EM & 0.327 & 1.653 & 2.883ms \\
    Ours & \textbf{0.296} & \textbf{1.559} & \textbf{2.175}ms  \\
    \bottomrule   
    \end{tabular}
    \vspace{-8pt}
\end{table}

\vspace{-3pt}
\subsection{Efficiency Verification}\vspace{-3pt}
The time cost of each module of our algorithm is shown in Tab.~\ref{tab3}. The time is mainly spent on Hybrid A* search, which is depended on the size and complexity of the environment map. The total time consumption of the algorithm is about hundreds milliseconds, which meets the requirements of real-time interception.
\begin{table}\vspace{-5pt}
    \centering
    \caption{Efficiency analyze}
    \vspace{-5pt}
	\label{tab3}
    \begin{tabular}{ccccc}
    \toprule    
     & Prediction & Path-Planner & Speed-Planner & Total  \\    
    \midrule   
    Time(Ave.) & $\leq$1ms & 150ms & $\leq$3ms & 154ms \\
    \bottomrule   
    \end{tabular}
\vspace{-10pt}
\end{table}

\vspace{-2pt}
\section{ Conclusion And Future Work}\label{conclusion}
\vspace{-5pt}
In this paper, we focus on the problem of intercepting the moving target with a wheeled mobile robot in a dynamic environment. We first predict the target's future trajectory through polynomial fitting based on the historical observation data. Then, we use Hybrid A* to search a coarse path and optimize it by gradient descent method. Finally we formulate the speed curve into piecewise B\'ezier curves optimization on ST graph, which is solved as a quadratic programming problem. The algorithm generates a feasible, effective and obstacle-avoidance trajectory for the wheeled robot while ensuring the accuracy of the interception.

Several directions may be explored as future work.  i) The moving target is no longer a single agent, but an organized group composed of multiple similar agents. ii) Consider the moving target no longer maintains a simple movement motion but has interaction with the environment, such as an escape strategy to the external interception.
\vspace{-5pt}
\bibliographystyle{IEEEtran}
\bibliography{reference}  % sigproc.bib is the name of the Bibliography in this case

\end{document}